\documentclass{article}

\PassOptionsToPackage{numbers, compress}{natbib}

\usepackage[preprint]{cml4impact_2022}

\usepackage[utf8]{inputenc} 
\usepackage[T1]{fontenc}    
\usepackage{hyperref}       
\usepackage{url}            
\usepackage{booktabs}       
\usepackage{amsfonts}       
\usepackage{nicefrac}       
\usepackage{microtype}      
\usepackage{xcolor}         

\usepackage{algorithm}
\usepackage{algorithmic}

\usepackage{nicefrac}
\usepackage{microtype}
\usepackage{graphicx}
\usepackage{caption}
\usepackage{subcaption}
\usepackage{tikz-cd}
\usetikzlibrary{arrows}
\usepackage{comment}
\usepackage{amssymb}
\usepackage{tikz}
\usepackage{tikz-qtree,tikz-qtree-compat}
\usetikzlibrary{positioning}

\usepackage{amsmath}
\usepackage{amsthm}
\usepackage{tikz}
\usepackage{tikz-qtree,tikz-qtree-compat}

\makeatletter
\newtheorem*{rep@theorem}{\rep@title}
\newcommand{\newreptheorem}[2]{%
\newenvironment{rep#1}[1]{%
 \def\rep@title{#2 \ref{##1}}%
 \begin{rep@theorem}}%
 {\end{rep@theorem}}}
\makeatother

\newtheorem{theorem}{Theorem}
\newreptheorem{theorem}{Theorem}

\usetikzlibrary{positioning}

\usepackage{stackengine}

\title{Unit Selection: Learning Benefit Function from Finite Population Data}

\author{%
  Ang Li, Song Jiang, Yizhou Sun, Judea Pearl\\
  Department of Computer Science\\
  University of California Los Angeles\\
  Los Angeles, CA 90095 \\
  \texttt{\{angli,songjiang,yzsun,judea\}@cs.ucla.edu} \\
}

\begin{document}

\maketitle

\begin{abstract}
The unit selection problem is to identify a group of individuals who are most likely to exhibit a desired mode of behavior, for example, selecting individuals who would respond one way if incentivized and a different way if not. The unit selection problem consists of evaluation and search subproblems. Li and Pearl defined the "benefit function" to evaluate the average payoff of selecting a certain individual with given characteristics. The search subproblem is then to design an algorithm to identify the characteristics that maximize the above benefit function. The hardness of the search subproblem arises due to the large number of characteristics available for each individual and the sparsity of the data available in each cell of characteristics. In this paper, we present a machine learning framework that uses the bounds of the benefit function that are estimable from the finite population data to learn the bounds of the benefit function for each cell of characteristics. Therefore, we could easily obtain the characteristics that maximize the benefit function.
\end{abstract}

\section{Introduction}
The unit selection problem frequently appears in the industries such as health science, political science, and marketing. In customer relationship management \cite{berson1999building, hung2006applying, lejeune2001measuring, tsai2009customer}, for example, companies are interested in discovering customers who are possible to leave but would not if there is an incentive. The incentive should be sent very carefully because the behavior of the customers is influenced by any incentives that the companies sent before. For another example, in online advertisement, companies are interested in identifying users who would view an advertisement if and only if the advertisement is prompted  \cite{bottou2013counterfactual, li2020training, li2014counterfactual, sun2015causal, yan2009much}. The challenge in identifying these individuals stems from the fact that the desired response pattern is not observed directly but rather is defined counterfactually in terms of what the individual would do under hypothetical unrealized conditions. For example, when a customer has bought a car with a considerable discount, we have no idea whether they would buy the same car if there is no such discount.

As defined by Li and Pearl \cite{li:pea19-r488}, the unit selection problem entails two sub-problems, evaluation and search. The evaluation problem is to develop an estimable objective function that, if maximized over the set of observed characteristics C (available for each individual), would ensure an optimal counterfactual behavior for the selected group. The search task is to develop a search algorithm to determine individuals based both on their observed characteristics and the objective function devised above. The hardness of the search task arises due to the large number of characteristics available for each individual and the sparsity of the data available in each cell of characteristics.

For the evaluation sub-problem, the benefit function for the unit selection problem was defined by Li and Pearl \cite{li:pea19-r488}, and it properly captures the nature of the desired behavior. By making the assumption that the treatment has no effect on the population-specific characteristics, Li and Pearl derived tight bounds of the benefit function using experimental and observational data. Li and Pearl \cite{li2022unit} then narrowed the bounds of the benefit function using covariates information and their causal structure inspired by Mueller, Li, and Pearl \cite{pearl:etal21-r505} and Dawid et al. \cite{dawid2017} that the bounds of probabilities of causation could be narrowed using covariates information. In addition, the unit selection problem with nonbinary treatment and effect was studied by Li and Pearl in \cite{li:pea-r516, li:pea22-r517}.

In this study, we focus on the search sub-problem with binary treatment and effect. Nonbinary situations can be extended with \cite{li:pea-r516,li:pea22-r517,li2022bounds,zhang2022partial}.

Consider the popular motivating scenario by Li and Pearl \cite{li:pea19-r488}: a mobile carrier is interested in identifying customers who are about to churn in the future based on their $15$ observed characteristics such as gender, income, age, monthly payments, and so on. A considerable discount will then be offered to identified customers to encourage them to continue the service so as to increase their service renewal rate. The management has to be careful that only those customers who would continue their service if and only if they receive the offer can receive the discount, and all other types of customers cannot. The management then collected the observational and experimental data of their customers (we call this population data because the data are randomized over all customers. It is hard to collect data for each set of characteristics because 1) the total number of the set of characteristics is large, $2^{15}$; 2) some set of characteristics is rare to appear).

The simplest solution is to apply Li-Pearl's benefit function with benefit vector $(1,-1,-1,-1)$ or $(1, -1, -1, -2)$ to every set of characteristics and selects all sets such that the benefit function is positive. However, we have $2^{15}$ sets of characteristics, and according to Li and Pearl \cite{li:pea22-r518}, each set of characteristics requires roughly $1500$ experimental and $1500$ observational samples to have relatively precise estimations. Therefore, at least $49152000$ experimental and $49152000$ observational samples are needed to evaluate the benefit function of every set of characteristics. In addition, the experimental and observational data are collected from the entire population (i.e., all customers); therefore, some of the characteristics are rare to appear. Therefore, it is impractical to evaluate every set of characteristics given the finite population data. Li et al. \cite{li:pea22-r519} proposed that a machine learning model can learn the bounds of probabilities of causation. Similar work can be applied to the benefit function. In this paper, we believe that the behavior of the customers is determined by their characteristics; therefore, we present a machine learning framework that uses the bounds of the estimable benefit function as the label (of those sets of characteristics such that the benefit function has sufficient data to estimate), to compute the bounds of benefit function for every set of characteristics. 

\section{Preliminaries}
\label{related work}
In this section, we review Li and Pearl's benefit function of the unit selection problem \cite{li:pea19-r488}.

The language of counterfactuals in structural model semantics, as given in \cite{galles1998axiomatic,halpern2000axiomatizing}, is used in this paper. The basic counterfactual sentence ``Variable $Y$ would have the value $y$, had $X$ been $x$" is denoted by $Y_x=y$. We distinguish between experimental and observational data; the experimental data is in the form of the causal effects denoted by $P(y_x)$, and the observational data is in the form of joint probability denoted by $P(x, y)$. Without further specification, we use $X$ to denote treatment and $Y$ to denote effect. For simplicity purposes, $Y_x=y, Y_{x'}=y, Y_{x}=y', Y_{x'}=y'$ are denoted by $y_x, y_{x'}, y'_x, y'_{x'}$, respectively.

Same as Li and Pearl \cite{li:pea19-r488}, individual behavior was classified into four response types: labeled complier, always-taker, never-taker, and defier. Suppose the benefit of selecting one individual in each category are $\beta, \gamma, \theta, \delta$ respectively (i.e., the benefit vector is $(\beta, \gamma, \theta, \delta)$). Li and Pearl defined the objective function of the unit selection problem as the average benefit gained per individual. Suppose $x$ and $x'$ are binary treatments, $y$ and $y'$ are binary outcomes, and $c$ are population-specific characteristics, the objective function (i.e., benefit function) is following (If the goal is to evaluate the average benefit gained per individual for a specific population $c$, $argmax_c$ can be dropped.):
\begin{eqnarray*}
\label{liobj}
argmax_c \text{ }\beta P(y_{x},y'_{x'}|c)+\gamma P(y_{x},y_{x'}|c) +\theta P(y'_{x},y'_{x'}|c)+\delta P(y'_{x},y_{x'}|c).
\end{eqnarray*}
Using a combination of experimental and observational data, Li and Pearl established the most general tight bounds on this benefit function as follow (which we refer to as Li-Pearl's Theorem in the rest of the paper). The only constraint is that the population-specific characteristics are not a descendant of the treatment.
\begin{theorem}
\label{thm1}
Given a causal diagram $G$ and distribution compatible with $G$, let $C$ be a set of variables that does not contain any descendant of $X$ in $G$, then the benefit function $f(c)=\beta P(y_x,y'_{x'}|c)+\gamma P(y_x,y_{x'}|c)+ \theta P(y'_x,y'_{x'}|c) + \delta P(y_{x'},y'_{x}|c)$ is bounded as follows:
\begin{eqnarray*}
W+\sigma U\le f(c) \le W+\sigma L\text{~~~~~~~~if }\sigma < 0,\\
W+\sigma L\le f(c) \le W+\sigma U\text{~~~~~~~~if }\sigma > 0,
\end{eqnarray*}
where $\sigma, W,L,U$ are given by,
\begin{eqnarray*}
&&\sigma = \beta - \gamma - \theta + \delta,\\
&&W=(\gamma -\delta)P(y_x|c)+\delta P(y_{x'}|c)+\theta P(y'_{x'}|c),\\
&&L=\max\left\{
\begin{array}{c}
0,\\
P(y_x|c)-P(y_{x'}|c),\\
P(y|c)-P(y_{x'}|c),\\
P(y_x|c)-P(y|c)\\
\end{array}
\right\},\\
&&U=\min\left\{
\begin{array}{c}
P(y_x|c),\\
P(y'_{x'}|c),\\
P(y,x|c)+P(y',x'|c),\\
P(y_x|c)-P(y_{x'}|c)+\\+P(y,x'|c)+P(y',x|c)
\end{array}
\right\}.
\end{eqnarray*}
\end{theorem}

\section{Causal Model}
\label{causalmodel}
Similar to the probabilities of causation in \cite{li:pea22-r519}, the learned bounds of the benefit function need to compare to the true bounds. The data generating model should be stated explicitly. We used the second causal model in \cite{li:pea22-r518} (see the appendix for the detail). The model is depicted in Figure \ref{causal1}. 
$X$ is a binary treatment, $Y$ is a binary effect, and $Z$ is a set of $20$ independent binary characteristics (say $Z_1,...,Z_{20}$). The structural equations are as follow (for simplicity reason, we let $x=1,x'=0$, and $y=1,y'=0$):
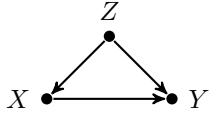
\begin{figure}
            \centering
            \begin{tikzpicture}[->,>=stealth',node distance=2cm,
              thick,main node/.style={circle,fill,inner sep=1.5pt}]
              \node[main node] (1) [label=above:{$Z$}]{};
              \node[main node] (3) [below left =1cm of 1,label=left:$X$]{};
              \node[main node] (4) [below right =1cm of 1,label=right:$Y$] {};
              \path[every node/.style={font=\sffamily\small}]
                (1) edge node {} (3)
                (1) edge node {} (4)
                (3) edge node {} (4);
            \end{tikzpicture}
            \caption{The Causal Model, where $X$ is a binary treatment, $Y$ is a binary effect, and $Z$ is a set of $20$ independent binary characteristics.}
            \label{causal1}
\end{figure}

\begin{eqnarray*}
Z_i &=& U_{Z_i} \text{ for } i \in \{1,...,20\},\\
X&=&f_X(M_X,U_X) =
\left \{
\begin{array}{cc}
1& \text{ if } M_X +U_X > 0.5, \\
0& \text{otherwize},
\end{array}
\right \}\\
Y&=&f_Y(X,M_Y,U_Y) =
\left \{
\begin{array}{cc}
1& \text{ if } 0< CX+M_Y+U_Y < 1 \text{ or } 1 < CX+M_Y+U_Y < 2\\
0& \text{otherwize},
\end{array}
\right \}\\
&&\text{where, } U_{Z_i}, U_X, U_Y \text{ are binary exogenous variables with Bernoulli distributions,}\\ &&\text{C is a constant, and }M_X,M_Y\text{ are linear combinatations of }Z_{1},...,Z_{20}.
\end{eqnarray*}
The value of $C,M_X,M_Y$ and the distributions of $U_X,U_Y,U_{Z_1},...,U_{Z_{20}}$ for the model are provided in the appendix.
\section{Benefit Function}
The benefit vector that we will use in the simulated study is $(1,-1,-1,-2)$, which is a common one to encourage complier as well as avoid all other three response types. Therefore, the benefit function is 
\begin{eqnarray*}
f(c) = P(y_{x},y'_{x'}|c)- P(y_{x},y_{x'}|c) - P(y'_{x},y'_{x'}|c)-2 P(y'_{x},y_{x'}|c).
\end{eqnarray*}

\section{Data Generating Process}
Similar to Li and Pearl \cite{li:pea22-r519}, $15$ of the $20$ binary characteristics were made observable (i.e., $Z_1,...,Z_{15}$), and $5$ of them were made unobservable (i.e., $Z_{16},...,Z_{20}$). Therefore, we have $2^{15}$ observed sets of characteristics.
\subsection{Informer Data}
The informer data contains the true bounds of the benefit function of each set of characteristics. From the SCM we have, the value of $X$ is determined by $U_X$ and $M_x$ (denoted by $f_X(M_X,U_X)$) and the value of $Y$ is determined by $X$, $M_Y$, and $U_Y$ (denoted by $f_Y(X,M_Y,U_Y)$). We first compute the true value of the benefit function and the true experimental and observational distributions for any set of $20$ characteristics $z=(z_1,...,z_{20})$. Since $z$ is fixed; therefore, $M_X$ and $M_Y$ are fixed (denoted by $M_X(z)$ and $M_Y(z)$). We then have the benefit function $f(z)$, experimental distribution $P(Y|do(X),z)$, and observational distribution $P(Y|X,z)$ as follows:
\begin{eqnarray*}
f(z) &=& P(y_{x},y'_{x'}|z)- P(y_{x},y_{x'}|z) - P(y'_{x},y'_{x'}|z)-2 P(y'_{x},y_{x'}|z)\\
&=& P(U_Y=0)(T_0-T_2-T_4-2T_6) + P(U_Y=1)(T_1-T_3-T_5-2T_7), \\
\text{where}, &T_0& =
\left \{
\begin{array}{cc}
1& \text{ if } Y(0,M_Y(z),0)=0 \text{ and } Y(1,M_Y(z),0)=1, \\
0& \text{otherwize},
\end{array}
\right \},\\
&T_1& =
\left \{
\begin{array}{cc}
1& \text{ if } Y(0,M_Y(z),1)=0 \text{ and } Y(1,M_Y(z),1)=1, \\
0& \text{otherwize}
\end{array}
\right \},\\
&T_2& =
\left \{
\begin{array}{cc}
1& \text{ if } Y(0,M_Y(z),0)=1 \text{ and } Y(1,M_Y(z),0)=1, \\
0& \text{otherwize},
\end{array}
\right \},\\
&T_3& =
\left \{
\begin{array}{cc}
1& \text{ if } Y(0,M_Y(z),1)=1 \text{ and } Y(1,M_Y(z),1)=1, \\
0& \text{otherwize}
\end{array}
\right \},\\
&T_4& =
\left \{
\begin{array}{cc}
1& \text{ if } Y(0,M_Y(z),0)=0 \text{ and } Y(1,M_Y(z),0)=0, \\
0& \text{otherwize},
\end{array}
\right \},\\
&T_5& =
\left \{
\begin{array}{cc}
1& \text{ if } Y(0,M_Y(z),1)=0 \text{ and } Y(1,M_Y(z),1)=0, \\
0& \text{otherwize}
\end{array}
\right \},\\
&T_6& =
\left \{
\begin{array}{cc}
1& \text{ if } Y(0,M_Y(z),0)=1 \text{ and } Y(1,M_Y(z),0)=0, \\
0& \text{otherwize},
\end{array}
\right \},\\
&T_7& =
\left \{
\begin{array}{cc}
1& \text{ if } Y(0,M_Y(z),1)=1 \text{ and } Y(1,M_Y(z),1)=0, \\
0& \text{otherwize}
\end{array}
\right \}.
\end{eqnarray*}
\begin{eqnarray*}
&&P(Y=1|do(X),z)\\
&=& P(U_Y=0)*Y(X,M_Y(z),0) + P(U_Y=1)*Y(X,M_Y(z),1).
\end{eqnarray*}
\begin{eqnarray*}
&&P(Y=1|X,z)\\
&=& P(U_X=0)*P(U_Y=0)*Y(X(M_X(z),0),M_Y(z),0)+\\
&&P(U_X=0)*P(U_Y=1)*Y(X(M_X(z),0),M_Y(z),1)) +\\ &&P(U_X=1)*P(U_Y=0)*Y(X(M_X(z),1),M_Y(z),0)) +\\ &&P(U_X=1)*P(U_Y=1)*Y(X(M_X(z),1),M_Y(z),1)).
\end{eqnarray*}
Second, we need to compute the informer data for any $15$ observed set of characteristics, say $c=(z_1,...,z_{15})$. $c=(z_1,...,z_{15})$ consists $32$ sets of $20$ characteristics (say $s_{0}=(z_1,...,z_{15},0,0,0,0,0), s_{1}=(z_1,...,z_{15},0,0,0,0,1), s_{2}=(z_1,...,z_{15},0,0,0,1,0), ...,s_{31}=(z_1,...,z_{15},1,1,1,1,1)$), then we have the benefit function $f(c)$, experimental distribution $P(Y|do(X),c)$, and observational distribution $P(Y|X,c)$ of any observed sets of characteristics $c$ are as follow:
\begin{eqnarray*}
f(c) &=& P(Y=0_{X=0}, Y=1_{X=1}|c)\\
&=& P(s_{0})/P(c)f(s_{0})+P(s_{1})/P(c)f(s_{1})+P(s_{2})/P(c)f(s_{2})+...+P(s_{31})/P(c)f(s_{31})\\
&=& P(Z_{16}=0)P(Z_{17}=0)P(Z_{18}=0)P(Z_{19}=0)P(Z_{20}=0)f(s_{0})+\\
&&P(Z_{16}=0)P(Z_{17}=0)P(Z_{18}=0)P(Z_{19}=0)P(Z_{20}=1)f(s_{1})+...+\\
&&P(Z_{16}=1)P(Z_{17}=1)P(Z_{18}=1)P(Z_{19}=1)P(Z_{20}=1)f(s_{31}).
\end{eqnarray*}
\begin{eqnarray*}
&&P(Y=1|do(X),c)\\
&=& P(Z_{16}=0)P(Z_{17}=0)P(Z_{18}=0)P(Z_{19}=0)P(Z_{20}=0)P(Y=1|do(X),s_{0})+\\
&& P(Z_{16}=0)P(Z_{17}=0)P(Z_{18}=0)P(Z_{19}=0)P(Z_{20}=1)P(Y=1|do(X),s_{1})+...+\\
&& P(Z_{16}=1)P(Z_{17}=1)P(Z_{18}=1)P(Z_{19}=1)P(Z_{20}=1)P(Y=1|do(X),s_{31}).
\end{eqnarray*}
\begin{eqnarray*}
&&P(Y=1|X,c)\\
&=& P(Z_{16}=0)P(Z_{17}=0)P(Z_{18}=0)P(Z_{19}=0)P(Z_{20}=0)P(Y=1|X,s_{0})+\\
&& P(Z_{16}=0)P(Z_{17}=0)P(Z_{18}=0)P(Z_{19}=0)P(Z_{20}=1)P(Y=1|X,s_{1})+...+\\
&& P(Z_{16}=1)P(Z_{17}=1)P(Z_{18}=1)P(Z_{19}=1)P(Z_{20}=1)P(Y=1|X,s_{31}).
\end{eqnarray*}
Therefore, the true bounds of the benefit function $f(c)$ for each set of observed characteristics could be obtained using Theorem \ref{thm1} and above observational and experimental distributions.

\subsection{Experimental Sample}
The $5000000$ finite experimental samples for training purposes can be obtained repeatedly as follows:
\begin{itemize}
    \item Randomly generate $(U_X,U_Y,U_{Z_1},...,U_{Z_{20}})$ using the given Bernoulli distributions;
    \item Randomly generate $X$ using $Bernoulli(0.5)$;
    \item Compute $Y=f_Y(X,M_Y,U_Y)$;
    \item Collect a experimental sample $(U_{Z_1},U_{Z_2},...,U_{Z_{15}},X,Y)$ (Note that $Z_i=U_{Z_{i}}$ in our model);
\end{itemize}

\subsection{Observational Sample}
The $5000000$ finite observational samples for training purposes can be obtained repeatedly as follows:
\begin{itemize}
    \item Randomly generate $(U_X,U_Y,U_{Z_1},...,U_{Z_{20}})$ using the given Bernoulli distributions;
    \item Compute $X=f_X(M_X,U_X)$;
    \item Compute $Y=f_Y(X,M_Y,U_Y)$;
    \item Collect a experimental sample $(U_{Z_1},U_{Z_2},...,U_{Z_{15}},X,Y)$ (Note that $Z_i=U_{Z_{i}}$ in our model);
\end{itemize}

\section{Machine Learning Model}
\subsection{Features and Label}
The training features are $15$ observed characteristics. The training labels are lower and upper bounds of the benefit function for the $15$ observed characteristics (two separate models for lower and upper bounds). In order to create the labels, we need Frequentist estimates of the experimental and observational distributions for that $15$ observed characteristics. According to Li, Mao, and Pearl \cite{li:pea22-r518}, $1300$ of experimental and observational samples are needed for the labels of the $15$ observed characteristics. Therefore, in our $5000000$ experimental and $5000000$ observational samples, if the same set of $15$ observed characteristics appeared more than $1300$ times, we obtain a lower bound and an upper bound for that set of $15$ observed characteristics by applying the Theorem \ref{thm1}. We have $302$ sets of characteristics for lower bound and $321$ sets of characteristics for upper bound for training purposes. The $61$ sets of lower bound and $65$ sets of upper bound are then split for the testing set.
\subsection{Learning}

In this paper, we use neural networks as our estimator to estimate the lower and upper bounds. Because the simulated data is tabular in nature, we just use a multilayer perceptron (MLP) in the following experiments. Our experimental environment is an AWS p3.2xlarge instance. The key parameters of our model are: embeddings dimension as 128, training epochs as 600, and learning rate as 0.01. 

\section{Experimental Results}
\begin{figure}
\centering
\begin{subfigure}[b]{0.49\textwidth}
\includegraphics[width=0.99\textwidth]{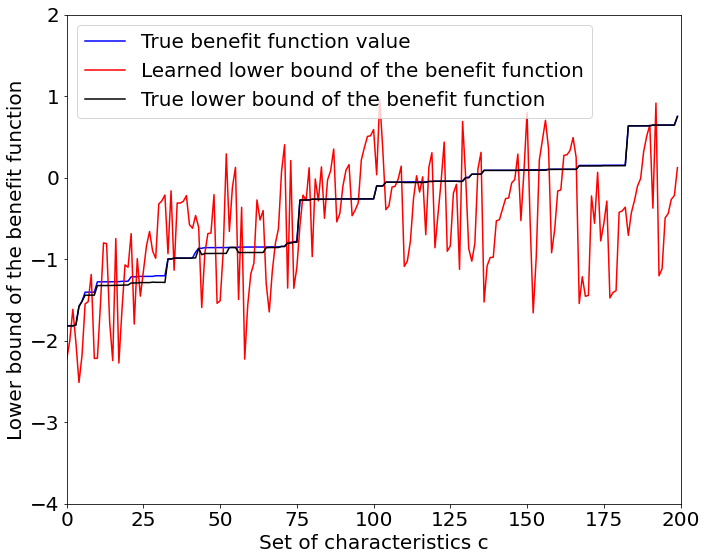}
\caption{Lower bound of the benefit function for $200$ sets of characteristics. These $200$ sets of characteristics are randomly selected from $32768$ of them.}
\label{res1}
\end{subfigure}
\hfill
\begin{subfigure}[b]{0.49\textwidth}
\includegraphics[width=0.99\textwidth]{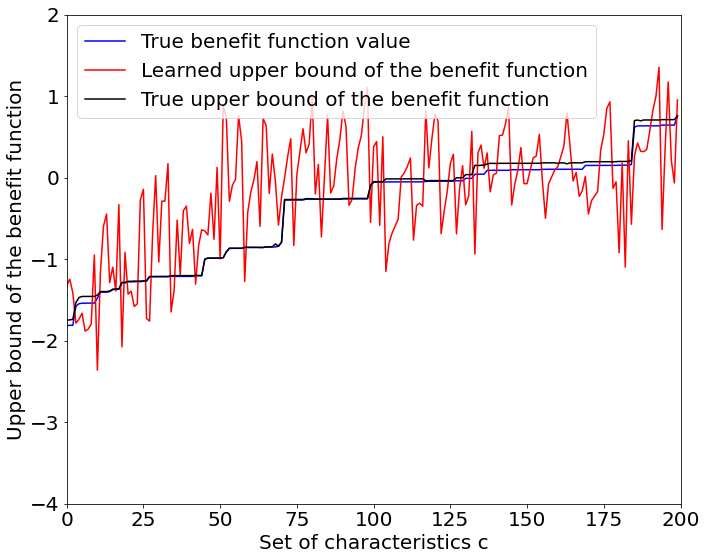}
\caption{Upper bound of the benefit function for $200$ sets of characteristics. These $200$ sets of characteristics are randomly selected from $32768$ of them.}
\label{res2}
\end{subfigure}
\caption{The learned bounds of the benefit function for the sets of characteristics compared to the true bounds of the benefit function.}
\label{f1}
\end{figure}

To show the performance of estimation, we randomly sampled $200$ sets of characteristics from all $32768$ instances. We observe that the average error of the learned lower bound and upper bound are $0.5652$ and $0.5447$, respectively. $\sim0.5$ is not of bad quality because the potential values of the benefit function are from $-4$ to $1$. We show how our predicted bounds fit the true ones in Figure~\ref{f1}, which confirms that the predicted bounds generally capture the theoretical ones.


\section{Discussion}
We demonstrated that the benefit function for all sets of characteristics could be learned from finite population data. We further discuss some properties of our proposed framework.

First, the bounds of the benefit function were successfully learned by the machine learning model. Similar to Li and Pearl in \cite{li:pea22-r519}, this is another example that the quantities of a counterfactual query could be learned properly if the proper counterfactual labels were fed to the machine learning model. This is the key to applying the machine learning model to causality concepts.

Second, we also applied the simplest machine learning model. The benefit vector and the data-generating process are available for researchers to apply fancy machine learning models. The accuracy of the predictions still has the potential to improve.

\section{Conclusion}
We illustrated how to learn the bounds of the benefit function in the unit selection problem for each set of characteristics using finite population data. All sets of characteristics of the bounds of the benefit function were being learned by a machine learning model, thus being able to select the sets of characteristics that have high (or positive) benefit values. Experiments showed that the benefit function is learnable with proper labels.

\section*{Acknowledgements}
This research was supported in parts by grants from the National Science
Foundation [\#IIS-2106908 and \#IIS-2231798], Office of Naval Research [\#N00014-21-1-2351], and Toyota Research Institute of North America
[\#PO-000897].
\bibliographystyle{plain}
\bibliography{neurips_2022.bib}

\newpage
\appendix

\section{Appendix}

\subsection{The Causal Model}
We used the second causal model in \cite{li:pea22-r518}. $M_X,M_Y$ and $C$ were uniformly generated from $[-1,1]$, and the Bernoulli parameters of $U_X,U_Y,U_{Z_1},...,U_{Z_{20}}$ were generated uniformly from $[0,1]$. The detailed model is as follows:
\begin{eqnarray*}
Z_i &=& U_{Z_i} \text{ for } i \in \{1,...,20\},\\
X&=&f_X(M_X,U_X) =
\left \{
\begin{array}{cc}
1& \text{ if } M_X +U_X > 0.5, \\
0& \text{otherwize},
\end{array}
\right \}\\
Y&=&f_Y(X,M_Y,U_Y) =
\left \{
\begin{array}{cc}
1& \text{ if } 0< CX+M_Y+U_Y < 1 \text{ or } 1 < CX+M_Y+U_Y < 2,\\
0& \text{otherwize},
\end{array}
\right \}\\
&&\text{where, } U_{Z_i}, U_X, U_Y \text{ are binary exogenous variables with Bernoulli distributions.}
s.t., \\
&&U_{Z_1} \sim \text{Bernoulli}(0.524110233482), U_{Z_2} \sim \text{Bernoulli}(0.689566064108),\\
&&U_{Z_3} \sim \text{Bernoulli}(0.180145428970), U_{Z_4} \sim \text{Bernoulli}(0.317153536644),\\
&&U_{Z_5} \sim \text{Bernoulli}(0.046268153873), U_{Z_6} \sim \text{Bernoulli}(0.340145244411),\\
&&U_{Z_7} \sim \text{Bernoulli}(0.100912238566), U_{Z_8} \sim \text{Bernoulli}(0.772038172066),\\
&&U_{Z_9} \sim \text{Bernoulli}(0.913108434869), U_{Z_{10}} \sim \text{Bernoulli}(0.364272299067),\\
&&U_{Z_{11}} \sim \text{Bernoulli}(0.063667554704), U_{Z_{12}} \sim \text{Bernoulli}(0.454839320009),\\
&&U_{Z_{13}} \sim \text{Bernoulli}(0.586687215140), U_{Z_{14}} \sim \text{Bernoulli}(0.018824647595),\\
&&U_{Z_{15}} \sim \text{Bernoulli}(0.871017316787), U_{Z_{16}} \sim \text{Bernoulli}(0.164966968157),\\
&&U_{Z_{17}} \sim \text{Bernoulli}(0.578925020078), U_{Z_{18}} \sim \text{Bernoulli}(0.983082980658),\\
&&U_{Z_{19}} \sim \text{Bernoulli}(0.018033993991), U_{Z_{20}} \sim \text{Bernoulli}(0.074629121266),\\
&&U_{X} \sim \text{Bernoulli}(0.29908139311), U_{Y} \sim \text{Bernoulli}(0.9226108109253),\\
&&C= 0.975140894243,
\end{eqnarray*}
\begin{eqnarray*}
&M_X& =
\begin{bmatrix}
Z_1~Z_2~...~Z_{20}
\end{bmatrix}\times
\begin{bmatrix}
0.843870221861\\ 0.178759296447\\ -0.372349746729\\ -0.950904544846\\ -0.439457721339\\ -0.725970103834\\ -0.791203963585\\ -0.843183562918\\ -0.68422616618\\ -0.782051030131\\ -0.434420454146\\ -0.445019418094\\ 0.751698021555\\ -0.185984172192\\ 0.191948271392\\ 0.401334543567\\ 0.331387702568\\ 0.522595634402\\ -0.928734581669\\ 0.203436441511
\end{bmatrix},
M_Y =
\begin{bmatrix}
Z_1~Z_2~...~Z_{20}
\end{bmatrix}\times
\begin{bmatrix}
-0.453251661832\\ 0.424563325534\\ 0.0924810605305\\ 0.312680246141\\ 0.7676961338\\ 0.124337421843\\ -0.435341306455\\ 0.248957751703\\ -0.161303883519\\ -0.537653062121\\ -0.222087991408\\ 0.190167775134\\ -0.788147770713\\ -0.593030174012\\ -0.308066297974\\ 0.218776507777\\ -0.751253645088\\ -0.11151455376\\ 0.785227235182\\ -0.568046522383
\end{bmatrix}
\end{eqnarray*}
\end{document}